\documentclass[conference]{IEEEtran}
\IEEEoverridecommandlockouts
\usepackage{cite}
\usepackage{amsmath,amssymb,amsfonts}
\usepackage{algorithmic}
\usepackage{graphicx}
\usepackage{textcomp}
\usepackage{xcolor}
\usepackage{hyperref}

\usepackage{algorithm}
\allowdisplaybreaks

\usepackage{accents}

\newcommand\numberthis{\addtocounter{equation}{1}\tag{\theequation}}
\def\BibTeX{{\rm B\kern-.05em{\sc i\kern-.025em b}\kern-.08em
    T\kern-.1667em\lower.7ex\hbox{E}\kern-.125emX}}
\begin{document}

\title{Optimal Vehicle Trajectory Planning for Static Obstacle Avoidance using Nonlinear Optimization
}

\author{Yajia Zhang* \and Hongyi Sun \and Ruizhi Chai \and Daike Kang \and Shan Li \and Liyun Li
\thanks{The authors are with Xmotors.ai, a subsidiary of Guangzhou Xiaopeng Motors Technology Co Ltd.
{\small 3940 Freedom Circle, Santa Clara, CA, USA 95054} 
*Corresponding author: Yajia Zhang 
{\small yajiaz@xmotors.ai} 
}
}


\maketitle

\begin{abstract}
Vehicle trajectory planning is a key component for an autonomous driving system. A practical system not only requires the component to compute a feasible trajectory, but also a comfortable one given certain comfort metrics. Nevertheless, computation efficiency is critical for the system to be deployed as a commercial product. In this paper, we present a novel trajectory planning algorithm based on nonlinear optimization. The algorithm computes a kinematically feasible and comfort-optimal trajectory that achieves collision avoidance with static obstacles. Furthermore, the algorithm is time efficient. It generates an 6-second trajectory within 10 milliseconds on an Intel i7 machine or 20 milliseconds on an Nvidia Drive Orin platform. 

\end{abstract}


\section{Introduction}
\label{sec:introduction}
The goal of vehicle trajectory planning is to find feasible trajectories for autonomous vehicles to navigate safely and comfortably in the environment. For urban driving, autonomous vehicles need the ability to travel through complex scenarios, e.g., narrow passages formed by parked vehicles. Furthermore, to account for environment or perception changes, the planning algorithm is normally required to generate trajectories at 10 hertz rate or even higher. Thus computation efficiency is a critical factor on evaluating a trajectory planning algorithm.      

\begin{figure}
    \centering
    \includegraphics[width=0.30\textwidth]{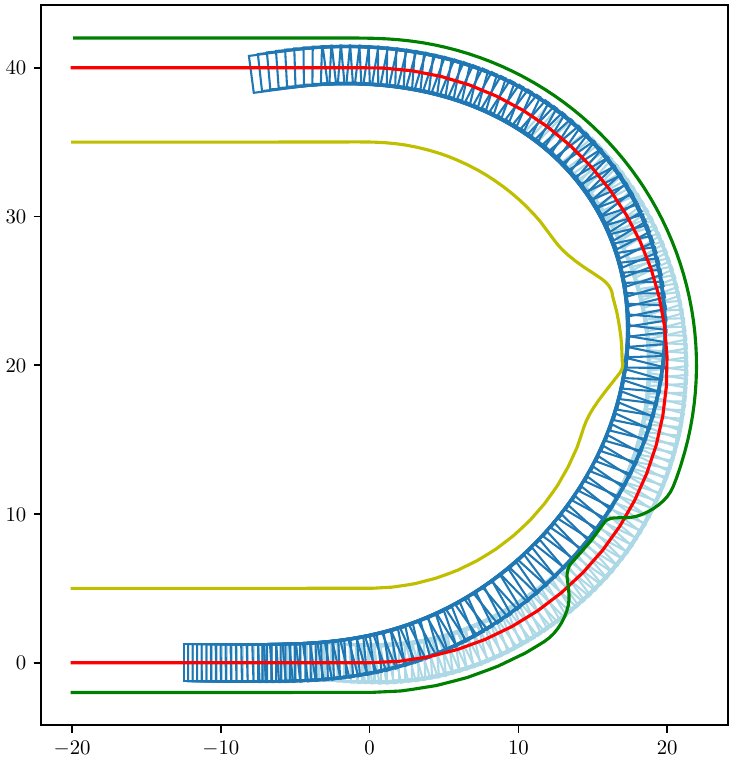}
    \caption{A $12$-second trajectory (in dark blue, with resolution $t = 0.1s$) generated by the proposed planning algorithm for a synthetic problem. Red line is the reference line, i.e, road centerline. Yellow and green lines are the left and right boundaries of the drivable corridor respectively. Note, the irregular shape of the corridor boundaries are due to incorporation of static obstacles. The trajectory in light blue is the initial guess for the optimization procedure.}
    \label{fig:trajectory}
\end{figure}

In this paper, we present a novel optimal trajectory planning algorithm for autonomous vehicles using nonlinear numerical optimization. Given a drivable corridor that implicitly incorporates the geometry of static obstacles, the trajectory planner computes a kinematically feasible trajectory within the drivable corridor with optimal comfort metrics (see Fig. \ref{fig:trajectory}). We adopt the methodology to hierarchically solve autonomous vehicle trajectory planning by planning path and speed profile in a sequential order. Path planning is responsible for providing the geometrical component of a trajectory to avoid collisions with static obstacles; speed profile planning assigns the temporal component given the planned path to avoid collisions with dynamic obstacles. This hierarchical approach breaks the trajectory planning problem into two sequential lower dimensional problems and greatly reduces the overall complexity. Though the proposed algorithm in this paper can be extended to avoid dynamic obstacles given their predicted trajectories, \textbf{we limit our application of the algorithm for static obstacle collision avoidance}. A latter step, a speed profile planner uses a similar algorithm in \cite{zhang2019nonlinear} to override the speed profile of the planned trajectory if necessary to avoid collisions with dynamic obstacles.

For autonomous vehicles, we normally have direct controls over acceleration through throttle and brake, and the angle of the steering wheel that corresponds to the curvature of a vehicle's trajectory according a vehicle bicycle model (see Fig. \ref{fig:bicycle-model}). In order to obtain graceful motions, these physical control commands must be continuous over time. This means we need to take linear jerk, i.e., $j = da/dt$ and curvature change rate, i.e., $\dot{\kappa} = d\kappa/dt$, which are at least one order higher than the control commands, into consideration in trajectory planning.

Compared to closed-form modelings of trajectories, discretized or piecewise modeling has the advantage on flexibility. To tackle complex urban driving scenarios, we model the trajectory using a piecewise function, with each piece corresponding to a constant ``control" variable pair $(j, \dot{\kappa})$ that transits from one vehicle state to the next. These sequence of control variable pairs decide the spatial and temporal information of the trajectory. Nevertheless, though our goal is to find an optimal control sequence essentially, it is difficult to use these control variables directly as optimization variables due to a chain-like dependency, i.e., the selection of $i$th control pair depends on all previous control pairs from $1$ to $i-1$. The chain-like dependency results a dense Jacobian and Hessian matrices in optimization and thus makes them computationally expensive to update at every iteration. To solve this, we use discretized vehicle states as optimization variables instead of control variables. A discretized vehicle state now only depends on its previous state and the dependency is sparse through all optimization variables. To satisfy kinematic constraints and maintain the connectivity of the trajectory, we also set up equality constraints between consecutive discretized states according to the kinematic model.  

Besides sparse dependency formulation, we also use a couple of techniques to speed up the nonlinear optimization procedure and improve the convergence rate, including smoothing drivable corridor boundaries, caching intermediate optimization results, etc. These techniques reduce at least $50\%$ of the computation time.



 


\section{Related Work}
\label{sec:related_work}
Trajectory planning for autonomous vehicles has been a vigorous research topic since the DARPA Grand and Urban Challenge. A number of algorithms have been developed to tackle the challenges \cite{5940562, miller2008team, urmson2008autonomous, kuwata2009real, thrun2006stanley}. These algorithms can be categorized into a couple of classes: randomized planners, such as Rapidly Exploring Random Tree (RRT)\cite{lavalle2001randomized}, C-PRM\cite{song2001randomized}, can be used for car-like robots with differential constraints. Some can even produce high-quality paths given enough computing time \cite{hwan2011anytime}. The problem is they are general planners intend to work regardless of the target problem and they lack the ability to exploit and utilize the dedicated road-like structured environment for autonomous vehicles, i.e., environment with directional centerline and boundaries. Thus, the computed trajectories are usually sub-optimal for autonomous vehicles, especially at high speed. The application of these algorithms are mostly limited to off-road scenarios, e.g., parking. 

Discrete search methods, such as \cite{frazzoli_primitive}, compute a trajectory by searching from a set of precomputed maneuvers and concatenating them to a maneuver sequence. The concatenation is performed by examining whether the starting state of one maneuver is sufficiently close to the ending state of the target maneuver. For simple environment with fewer obstacles, these methods generally work well. However, to deal with complex scenarios in urban driving, the number of precomputed maneuvers needs to grow exponentially, which makes these methods impractical. 


Instead of planning in the map frame, some methods \cite{werling2010optimal} transform the planning problem to a Frenet frame that is defined by a smooth driving guide line. The motion of the vehicle is decoupled to two 1-dimensional motions w.r.t. the driving guide line: longitudinal motion that moves along the driving guide line and lateral motions that moves perpendicularly to the driving guide line. Longitudinal and lateral motions are modeled using quintic or quartic polynomials and are planned individually. Then, longitudinal and lateral motions are combined and transformed back to map frame for constraint checking and selection. Our previous work \cite{Zhang_2020} adopts a similar idea in \cite{werling2010optimal} that plans a path/trajectory in a Frenet frame but we use piecewise trajectory modeling for the 1-dimensional motions to improve the flexibility. Frenet frame motion decoupling is advantageous since it can effectively exploit the road structure, and lower the complexity by reducing the dimensionality of planning. However, the drawbacks are also obvious: vehicle's kinematic constraints, such as curvature, curvature change rate, etc, are defined in the map frame and they become too complex in the Frenet frame (see Appendix in \cite{werling2010optimal}) to consider directly for planning; also the frame transformation requires a driving guide line that is $C^3$ continuous to accurately decouple the motion to the second-order derivative level (e.g., acceleration level), which requires a preprocessing of the driving guide line.
  
The most flexible methods that can deal with complex environments are optimization based methods. The work in \cite{ziegler2014making} runs a sequential quadratic programming procedure in the map frame to directly compute a trajectory. The trajectory is represented by a sequence of discretized positional points, and these positional points are used as optimization variables. The optimization procedure iteratively finds a sequence of points that minimizes the objective function that combines safety and comfort factors. The advantages of optimization methods in the map frame include direct modeling of trajectory optimality and enforcement of constraint satisfaction. Furthermore, as the trajectory is modeled as a sequence of densely discretized points, these methods achieve maximal control over the spatial and temporal of the trajectory to deal with complex scenarios.

Our proposed method follows a similar idea as in \cite{ziegler2014making} to use a discretized modeling of the trajectory and use these discretized points as optimization variables. But besides the positional information of one state, we use a populated state representation including heading, curvature, curvature change rate, etc, that are directly linked to kinematic feasibility. This augmented state representation makes the objective and constraints easier to formulate using exact form. Between consecutive states, a list of equality constraints are set up to maintain the kinematic feasibility and connectivity of the trajectory.    

In general, nonlinear optimization is more difficult to solve than quadratic programming. Some work such as \cite{cilqr} \cite{Zhang_2020} uses linear functions to linearize the nonlinear constraints and formulates the objective function into a quadratic form. However, the approximation sacrifices accuracy and could cause substantial deviations in modeling for the original problem. Our ideology is to keep the modeling close to the original problem, and use other techniques such as smoothing input functions, better guess of starting point, caching intermediate optimization results, etc, to speed up the overall optimization process.

\section{Problem Definition}
\label{sec:problem_definition}
\begin{figure}
    \centering
    \includegraphics[width=0.30\textwidth]{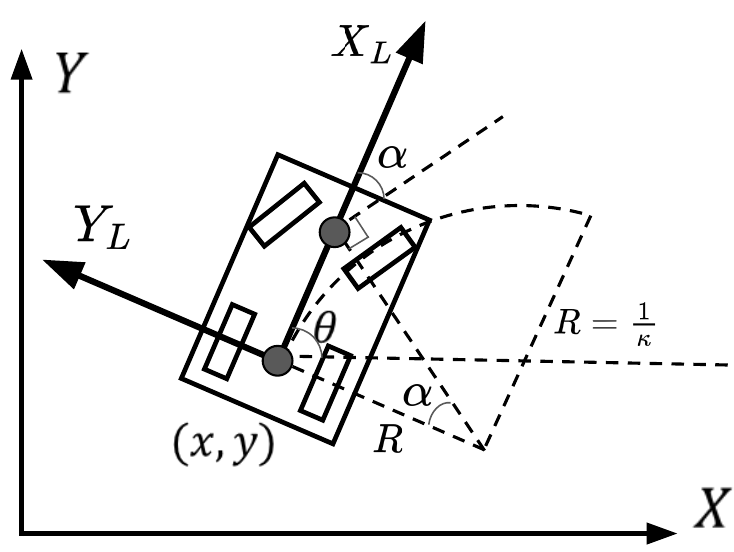}
    \caption{Illustration of a vehicle bicycle model. Bicycle model simplifies a four-wheeled vehicle to a two-wheeled bicycle with two wheels at the center of the front axis and rear axis respectively (dark points in the figure). The steering angle $\alpha$ leads to a circular movement of the vehicle with radius $R = L/\boldsymbol{\tan}(\alpha)$, where $L$ is the axis length, i.e., the distance between front and rear axis. $\kappa$ in the vehicle state space is the inverse of $R$, i.e., $\boldsymbol{\tan}(\alpha) / L$, and implicitly represents the steering angle of the vehicle.}
    \label{fig:bicycle-model}
\end{figure}

A state $s$ for an autonomous vehicle normally includes the following variables $(x, y, \theta, \kappa, v, a)$. $x$ and $y$ specify the coordinate of a reference point on the vehicle, normally the center of gravity or center of rear axis, in the map frame. $\theta$ is the heading angle of the vehicle in the map frame. $\kappa$ is the curvature of the turning circle given instant steering angle $\alpha$ (see Fig.\ref{fig:bicycle-model}). $v$ and $a$ are the linear velocity and acceleration respectively. The goal of trajectory planning is to find a function $\boldsymbol{S}(t)$ that maps a temporal parameter $t \in [0, t_{max}]$ to a feasible state $s$. 

\subsection{Vehicle Kinematic Model}
\label{subsec:kinematic_model}
Kinematic model specifies how the vehicle state evolve over time given a control input. We assume linear jerk $j$, i.e., $\frac{da}{dt}$, and curvature change rate $\dot{\kappa}$, i.e., $\frac{d\kappa}{dt}$, are control variables in the model. For a given time interval $\Delta t$, the transition from one vehicle state $s_{0}$ to the next state $s_{1}$ are given by the following equations, assuming $j$ and $\dot{\kappa}$ remain constant within time $\Delta t$:  

\begin{flalign*}
    a_{1} &= a_{0} + \boldsymbol{\Delta a}(\Delta t)  \numberthis \\
          &= a_{0} + \int_{0}^{\Delta t} \boldsymbol{j}(t) dt = a_{0} + j \Delta t   \\
    v_{1} &= v_{0} + \boldsymbol{\Delta v}(\Delta t) \numberthis \\
          &= v_{0} + \int_{0}^{\Delta t} \boldsymbol{a}(t) dt = v_{0} + a_{0} \Delta t + \frac{1}{2} j \Delta t^2   \\
    \kappa_{1} &= \kappa_{0} + \boldsymbol{\Delta \kappa}(\Delta t) \numberthis \\
        &= \kappa_{0} + \int_{0}^{\Delta t} \boldsymbol{\dot{\kappa}}(t) dt = \kappa_{0} + \dot{\kappa} \Delta t   \\
    \theta_{1} &= \theta_{0} + \boldsymbol{\Delta \theta}(\Delta t) \numberthis \\ 
    &= \theta_{0} + \int_{0}^{\Delta s} \boldsymbol{\kappa}(s) ds  \\
    &= \theta_{0} + \int_{0}^{\Delta t} \boldsymbol{\kappa}(\boldsymbol{s}(t)) \boldsymbol{s^{\prime}}(t) dt                                 \\
    &= \theta_{0} + \kappa_0 v_0 \Delta t + \frac{1}{2}(\kappa_0 a_0 + v_0 \dot{\kappa}) \Delta t^2 \\ &+ \frac{1}{3}(\frac{1}{2}\kappa_0 j + a_0 \dot{\kappa}) \Delta t^3 + \frac{1}{8} \dot{\kappa} j \Delta t^4   \\
    x_{1} &=x_{0} + \boldsymbol{\Delta x}(\Delta t) \numberthis \\
          &= x_{0} + \int_{0}^{\Delta s} \boldsymbol{cos}(\boldsymbol{\theta}(s)) ds   \\
            &= x_0 + \int_{0}^{\Delta t} \boldsymbol{cos}(\boldsymbol{\theta}(\boldsymbol{s}(t)))\boldsymbol{s^{\prime}}(t) dt  \\
    y_{1} &= y_{0} + \boldsymbol{\Delta y}(\Delta t) \numberthis \\
      &= y_{0} + \int_{0}^{\Delta s} \boldsymbol{sin}(\boldsymbol{\theta}(s)) ds  \\
            &= y_{0} + \int_{0}^{\Delta t} \boldsymbol{sin}(\boldsymbol{\theta}(\boldsymbol{s}(t)))\boldsymbol{s^{\prime}}(t) dt  \\
\end{flalign*}

$\boldsymbol{\Delta X}, X \in \{x, y, \theta, \kappa, v, a\}$ represents the increment function for each element in the state space according to the kinematic model. $s$ is an auxiliary variable representing the distance the vehicle has travelled. Though other functions are straightforward to evaluate, positional increment functions $\boldsymbol{\Delta x}$ and $\boldsymbol{\Delta y}$ are difficult to compute directly. The integrals in this form normally do not have closed-form solutions. To address the problem, we use Gaussian-Legendre quadrature to approximate the integration. The integration problem is then transformed to a weighted summation on target function values at given Gauss nodes: 

\begin{align*}
&\boldsymbol{\Delta x}(\Delta t) = \int_{0}^{\Delta t} \boldsymbol{cos}(\boldsymbol{\theta}(\boldsymbol{s}(t)))\boldsymbol{s^{\prime}}(t) dt \\
          &\approx \frac{1}{2} \Delta t \sum_{i = 1}^N w_i \boldsymbol{cos}(\boldsymbol{\theta}(\boldsymbol{s}(\frac{1}{2}\Delta t \xi_i + \frac{1}{2}\Delta t))) \boldsymbol{s}^\prime(\frac{1}{2}\Delta t \xi_i + \frac{1}{2}\Delta t) \\
&\boldsymbol{\Delta y}(\Delta t) = \int_{0}^{\Delta t} \boldsymbol{sin}(\boldsymbol{\theta}(\boldsymbol{s}(t)))\boldsymbol{s^{\prime}}(t) dt \\
          &\approx \frac{1}{2} \Delta t \sum_{i = 1}^N w_i \boldsymbol{sin}(\boldsymbol{\theta}(\boldsymbol{s}(\frac{1}{2}\Delta t \xi_i + \frac{1}{2}\Delta t))) \boldsymbol{s}^\prime(\frac{1}{2}\Delta t \xi_i + \frac{1}{2}\Delta t) \\
\end{align*}

$N$ is the order of the Gaussian-Legendre quadrature and it decides the level of accuracy the approximation can achieve. $\xi_i$ and $w_i$ are nodes and corresponding weights for the $N$th order Gauss-Legendre quadrature. In our implementation, we use $N = 10$.

\subsection{Trajectory Planning Problem Input}
\label{subsec:input}
The input to the proposed planner includes the following items:
\begin{itemize}
    \item Initial state $s_{0}$.
    \item Geometrical parameters of the vehicle, including length $l$, width $w$, axis length $l_{axis}$, for collision avoidance modeling.
    \item Vehicle's kinematic bounds, including maximal steering angle $\alpha_{max}$, maximal steering angle rate $\dot{\alpha}_{max}$, acceleration and jerk bounds, etc.
    \item Vehicle kinematic model (see Fig. \ref{fig:bicycle-model})
    \item Reference line $l_{ref}$, normally the centerline of the road, in the form of a directed polyline, i.e., a sequence of two dimensional points.
    \item Drivable corridor (see Fig. \ref{fig:collision_avoidance}) which consists two directed polylines formed by road boundaries and surrounding static obstacles.
    \item Target speed $v_{target}$ which is the desired speed of the vehicle, either from user input or speed limit from road.
\end{itemize}

\section{Optimization Formulation}
The trajectory planning problem is formulated as a numerical optimization problem. The constraints and objective function are nonlinear functions of the optimization variables, which makes the problem a typical nonlinear optimization problem.

\subsection{Trajectory Modeling and Variables}
The trajectory $\boldsymbol{S}(t)$ is modeled as a piecewise function. It is discretized by the temporal parameter $t$ with $\Delta t$ as resolution. Each discretized point corresponds to a vehicle state. 

\begin{center}
\label{tab:variables}
\begin{tabular}{c c c c c c c c c}
$x_0$       &                           &$x_1$          &                           &$x_2$      &           &$x_{n-2}$          &                           &$x_{n-1}$          \\ 
$y_0$       &                           &$y_1$          &                           &$y_2$      &           &$y_{n-2}$          &                           &$y_{n-1}$          \\   
$\theta_0$  & $\xrightarrow{\Delta t}$  &$\theta_1$     & $\xrightarrow{\Delta t}$  &$\theta_2$ &$\ldots$   &$\theta_{n-2}$     &$\xrightarrow{\Delta t}$   &$\theta_{n-1}$     \\
$\kappa_0$  &                           &$\kappa_1$     &                           &$\kappa_2$ &           &$\kappa_{n-2}$     &                           &$\kappa_{n-1}$     \\
$v_0$       &                           &$v_1$          &                           &$v_2$      &           &$v_{n-2}$          &                           &$v_{n-1}$          \\
$a_0$       &                           &$a_1$          &                           &$a_2$      &           &$a_{n-2}$          &                           &$a_{n-1}$          \\
\end{tabular}
\end{center}

Between consecutive states, we assume a constant linear jerk term $j$, and a constant third-order angular term $\dot{\kappa}$ as transition variables. To maintain the continuity of the trajectory, consecutive states must satisfy the equality relation defined by the kinematic model discussed in \ref{subsec:warm_start}. The optimality evaluation and constraint satisfaction checking of the trajectory are performed on these discretized trajectory points. In optimization, $(x, y, \theta, \kappa, v, a)$ at the discretized states, along with transition variables $j$ and $\dot{\kappa}$, are optimization variables. For a trajectory with $n$ discrete states, we have $8(n - 1)$ variables (excluding the initial state $s_0$). 

\subsection{Objective Formulation}
Optimality evaluation of a vehicle trajectory includes the following cost factors, and the objective function is a weighted summation of these costs:
\begin{itemize}
    \item \textbf{Centripetal acceleration} This factor penalizes the centripetal acceleration so that the autonomous vehicle needs to slow down at curves or cut straight at curvy roads. \\ 
    $f_{centri\_acc} = \Sigma (v_i^2 \kappa_i)^2$. 
    \item \textbf{Centripetal jerk} This factor penalizes the sudden of vehicles steering wheel when the speed is high. \\ 
    $f_{centri\_jerk} = \Sigma (2 v_i a_i \kappa_i + v_i^2 \dot{\kappa}_i)^2$
    \item \textbf{Curvature change rate} This factor penalizes the general sudden changes of steering wheel. \\
    $f_{\dot{\kappa}} = \Sigma \dot{\kappa}_i^2$ 
    \item \textbf{Linear jerk} This factor penalizes sudden accelerations and brakes. \\
    $f_{linear\_jerk} = \Sigma j_i^2$ 
    \item \textbf{Distance to the reference line} This factor encourages the vehicle to drive close to the center of the road. \\
    $f_{lat\_distance} = \Sigma \boldsymbol{L}_{\delta^l} (\boldsymbol{d}((x_i, y_i), \boldsymbol{P}_{\{x,y\}}(x_i, y_i, l_{ref})))$
    \item \textbf{Closeness to target speed}. To encourage the trajectory to travel further, we penalize the difference between vehicle's speed and a user-defined target speed. \\
    $f_{\Delta v} = \Sigma \boldsymbol{L}_{\delta^v} (v_i - v_{target})$
\end{itemize}    

$\Sigma$ represents the summation operation over state index $i \in [1, n - 1]$. $\boldsymbol{P}$ is a projection function that finds the projection point $(x_p, y_p)$, along its direction $\theta_p$, on a polyline $l$ given a point $(x, y)$. $\boldsymbol{d}$ is a function that computes the Euclidean distance between two points. $\boldsymbol{L}_\delta$ is a Huber loss function \cite{huber_loss}, which is commonly used in robust regression. We use Huber loss to model the cost of distance to reference line and closeness to target speed. In contrast to other cost factors in quadratic form, Huber loss function has the property that the cost is of quadratic form if the input is within a user defined $\delta$, linear if the input is outside. Thus it avoids the underlying cost factor from dominating the overall objective function when outlier input happens. Take the cost term of distance to reference line as an example, when autonomous vehicle initiates a lane change, a natural approach is to switch the reference line from ego lane (the lane where the ego vehicle is in) to the target lane. If a quadratic form of the cost is used, at the beginning of the lane change, this cost factor will become excessively large numerically, which causes other cost factors essentially become ineffective and leads to drastic lateral motions. The same idea applies to the closeness to target speed factor during ego vehicle starts from a static state to avoid drastic longitudinal motions.



\begin{figure}
    \centering
    \includegraphics[width=0.35\textwidth]{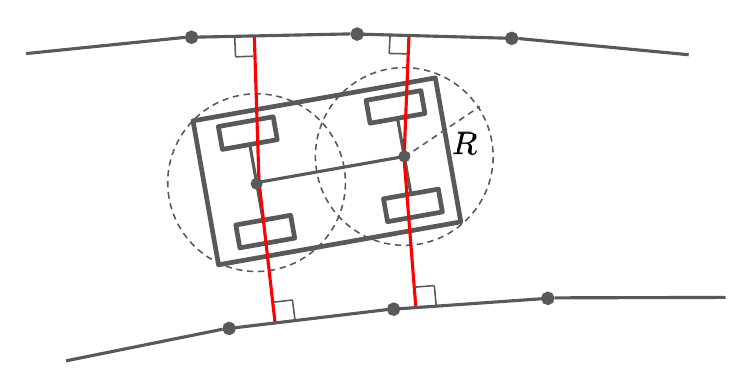}
    \caption{Illustration of drivable corridor boundaries and vehicle geometrical modeling for collision checking. Two circles with radius $R$ are constructed with origins at the middle of front and rear axises respectively. For vehicles with different geometry, the number of circles, the selection of origins and radius can be different to better fit a specific shape. Collision checking is performed by comparing the projection distances (in red) of two origins on to the drivable corridor boundaries and radius $R$.}
    \label{fig:collision_avoidance}
\end{figure}

\subsection{Constraints}
\subsubsection{Equality constraints for trajectory continuity}
To maintain the continuity of the planned trajectory, the following equality constraints derived from the kinematic model must be satisfied between consecutive states: 
\begin{align}
    X_{i + 1} - X_{i} - \boldsymbol{\Delta X}(\Delta t) = 0, X \in \{x, y, \theta, \kappa, v, a\}
\end{align}


\subsubsection{Collision avoidance constraints}
We use two circles with origins at the middle of the vehicle's front and rear axises to approximate the shape of the vehicle (see Fig. \ref{fig:collision_avoidance}). The same projection function $\boldsymbol{P}$ is used to project the origins of the circles onto the boundaries of the drivable corridor. The Euclidean distances between the origins and the projected points are used to check whether the vehicle is in collision with the boundaries.  

\subsubsection{Kinematic limit and pose constraints}
At each discretized state $s_i$, the following constraints must be satisfied:
\begin{itemize}
    \item Linear acceleration $a_i \in [a_{min}, a_{max}]$
    \item Linear jerk $j_i \in [j_{min}, j_{max}]$
    \item Linear velocity $v_i \in [0.0, +\infty)$
    \item Curvature $|\kappa_i| \leq \kappa_{max}$
    \item Centripetal acceleration $|v_i^2 \kappa_i| \leq a_{centri\_max}$
    \item Centripetal jerk $|2v_i a_i \kappa_i + v_i^2 \dot{\kappa}_i| \leq j_{centri\_max}$
    \item Heading angle difference \\
    $|\theta_i - \boldsymbol{P}_{\theta}(x_i, y_i, l_{ref})| \leq \theta_{diff\_max}$
\end{itemize}

The first six constraints are from vehicle's kinematic limits. The heading difference constraint is not necessary but we find it is helpful on convergence rate in large curvature scenarios such as U-turns as it reduces the volume of the search space.


\begin{figure}
    \includegraphics[width=0.48\textwidth]{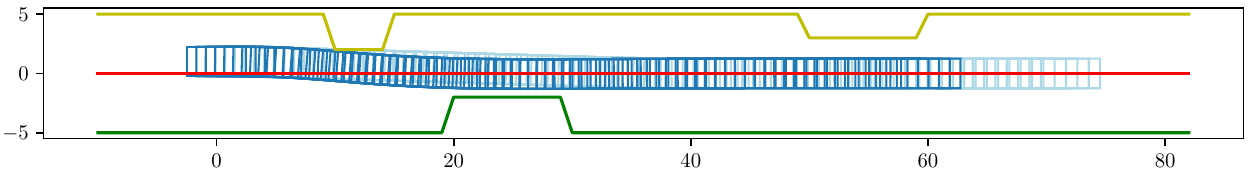} \\
    \includegraphics[width=0.48\textwidth]{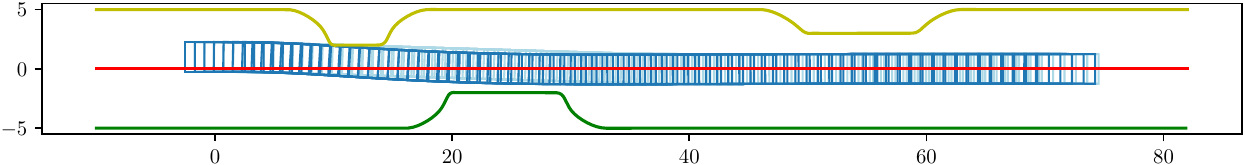}
    \caption{Comparison between original drivable corridor boundaries (top) and smoothed ones (bottom). The sharp corners are smoothed using a quadratic programming method described in \cite{Zhang_2020}. The optimizer takes 60 and 7 iterations to converge to an optimal trajectory respectively given original and smoothed corridor boundaries.}
    \label{fig:boundary_smoothing}
\end{figure}

\section{Implementation and Experiments}
We use optimization package IPOPT \cite{wachter2006implementation} with linear solver MA97 from HSL\cite{hslma97} to implement the proposed algorithm. We test the algorithm on a list of synthetic problems and conduct extensive road tests. The reported computation times are from an Intel i7 machine with 64GB memory for synthetic problems and an Nvidia Drive Orin platform for road tests. Vehicle geometry and kinematic bounds are set according to a specific vehicle model from \href{https://heyxpeng.com/}{Xiaopeng Motors}. Target speed $v_{target}$ is set dynamically according to road speed limit on road tests.

\subsection{Initial Guess of Trajectory and Warm Start}
\label{subsec:warm_start}
\begin{figure}
    \centering
    \includegraphics[width=0.45\textwidth]{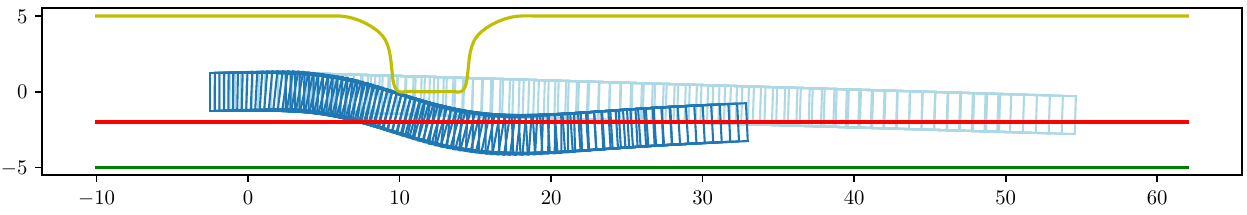}
    \includegraphics[width=0.48\textwidth]{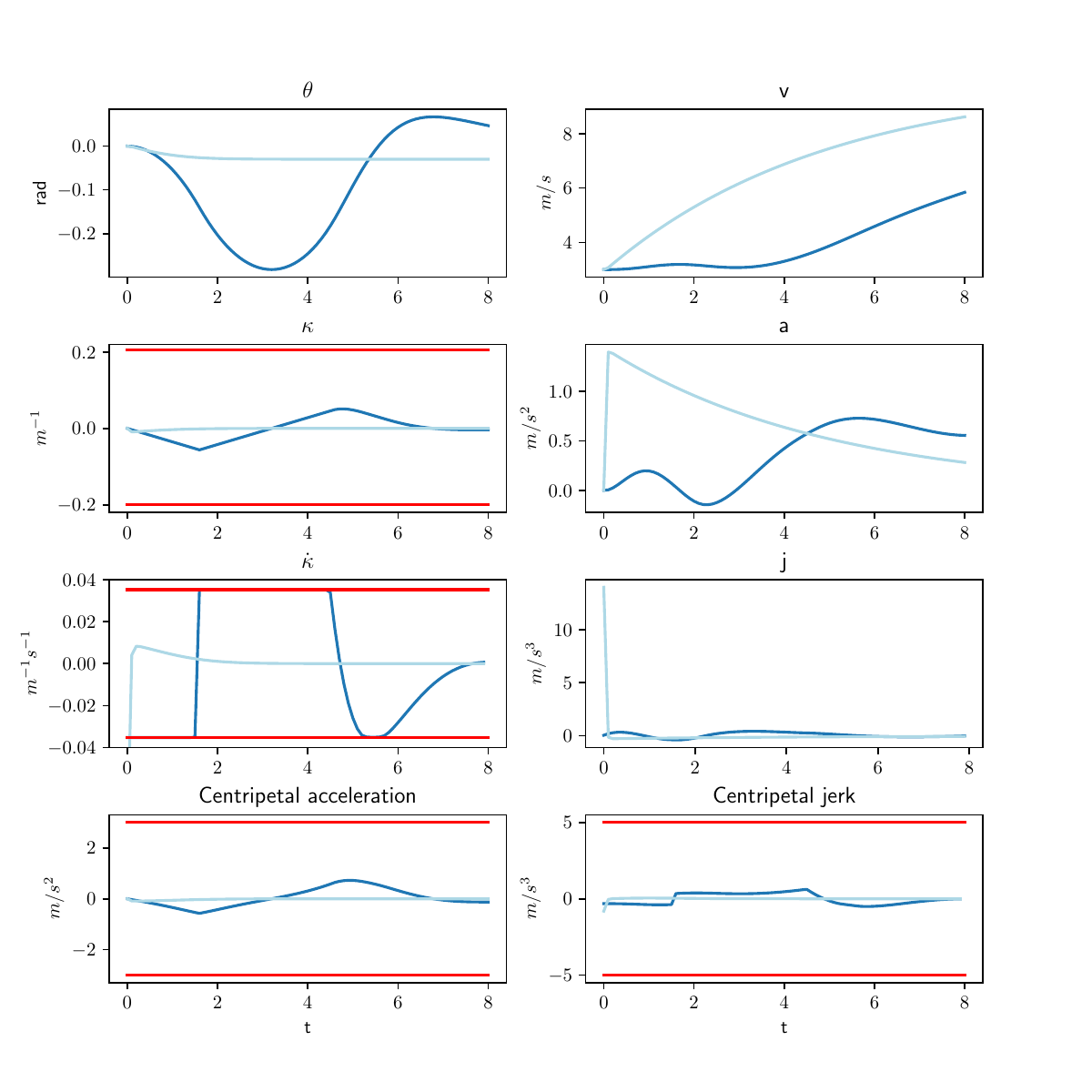}
    \caption{Planned trajectory and its geometrical and temporal properties (heading $\theta$, curvature $\kappa$, curvature temporal change rate $\dot{\kappa}$, velocity $v$, acceleration $a$ and linear jerk $j$) from a synthetic obstacle avoidance scenario. Computed centripetal acceleration and jerk are also shown. The limits of these properties are drawn in red. The planned trajectory is shown in dark blue and the initial guess from P-controller is shown in light blue. The trajectory is 8-second long. The autonomous vehicle is set to start from a moderate speed ($3m/s$). To avoid the collision with the static obstacle and maintain a comfort lateral motion, the vehicle needs to slow down its speed while making the lateral movement. This example takes the algorithm 46 iterations (computation time $89ms$) to converge to an optimal solution. Note, the initial guess is in collision with the boundary which makes the problem difficult to solve.}
    \label{fig:scenario_straight_road_obs}
\end{figure}

Initial guess is important in nonlinear optimization. We use a simple proportional controller to compute a trajectory for the optimizer to start with. The controller computes the longitudinal and lateral feedback gains based on the difference between vehicle's speed to target speed $v_{target}$ and lateral distance from vehicle's position to the given reference line $l_{ref}$ (see Fig.\ref{fig:scenario_straight_road_obs} for an example trajectory from the P-controller).

Furthermore, in road tests, if the planning for previous cycle is successful and the problem is similar to current one, e.g., the target reference lines are the same, the variable values along with their bound multipliers from previous cycle are used to ``warm start" the IPOPT solver. Our experiments show that in most cases ($> 90\%$), using warm start data can reduce the number of iterations to 2 or 3. 


    

\subsection{Smoothness of Drivable Corridor Boundaries}
We find the smoothness and density of the drivable corridor boundaries greatly affect the convergence rate of the optimization process, especially when the boundaries contain sharp corners due to the incorporation of static obstacles. Our intuitive explanation is that dense resolution of the drivable corridor (and reference line) can reduce projection error from the polyline projection function $\boldsymbol{P}$, and smoother edge can lead to smoother derivative changes. In our implementation, we perform a simple preprocessing step to smooth out sharp corners and linearly interpolate the boundaries to a higher resolution (from $1m$ to $0.1m$) for corridor boundaries (see Fig.\ref{fig:boundary_smoothing}).


\subsection{Implementation Optimization}
We profile the computation time of the algorithm and identify the time consuming bottlenecks. One bottleneck is the projection function $\boldsymbol{P}$ that projects a positional point $(x, y)$ onto a polyline (the reference line or drivable corridor boundaries in our case). One naive implementation of $\boldsymbol{P}$ is to loop through all the points in the polyline and locate the nearest line segment for projection, which is a $O(n)$ algorithm where $n$ is the number of point in the polyline. As IPOPT uses a filtered line search strategy, the projection will be called at each trial point and the computation time becomes significant. Our time profiling shows that the projection function alone takes over $40\%$ of overall computation time.

To improve this, we use R-tree \cite{guttman1984r} to index the polyline before the optimization procedure, and the projection algorithm becomes $O(log(n))$. Though there is an overhead time for indexing, this reduces $70\%$ of the time used for the projection function.  



\subsection{Testing Results}
\begin{figure}
    \centering
    \includegraphics[width=0.45\textwidth]{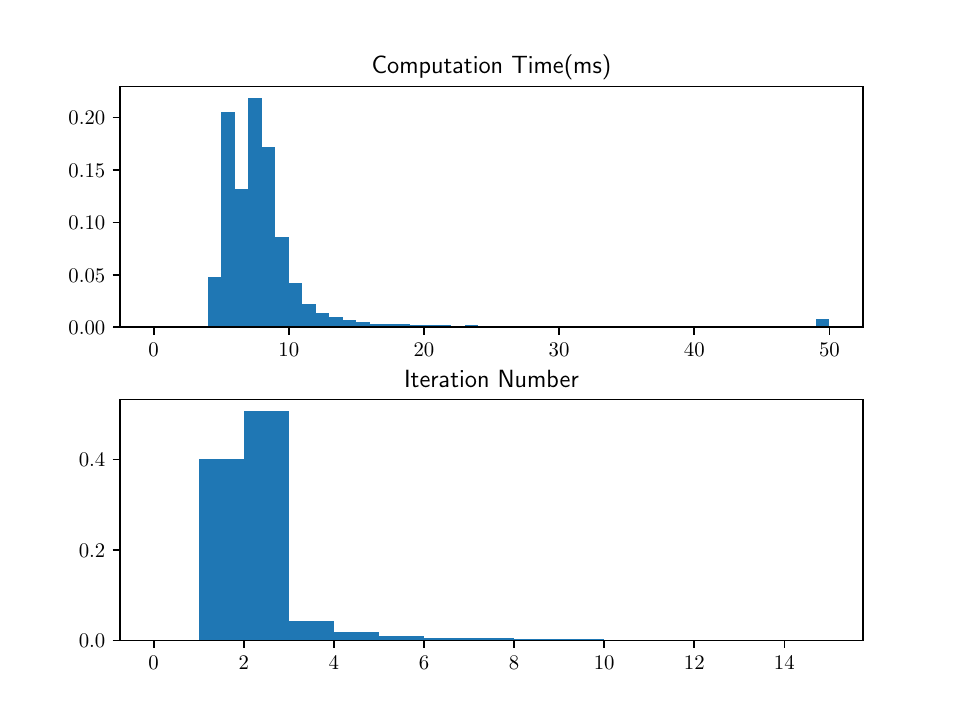}
    \caption{Histograms of computation time and number of optimization iterations from an over one hour test drive. The cutoff time for computation is $50$ ms. There are less than $1\%$ of the planning cycles that the algorithm fails to compute a feasible trajectory.}
    \label{fig:computation_time}
\end{figure}

\begin{figure}
    \centering
    \includegraphics[width=0.28\textwidth]{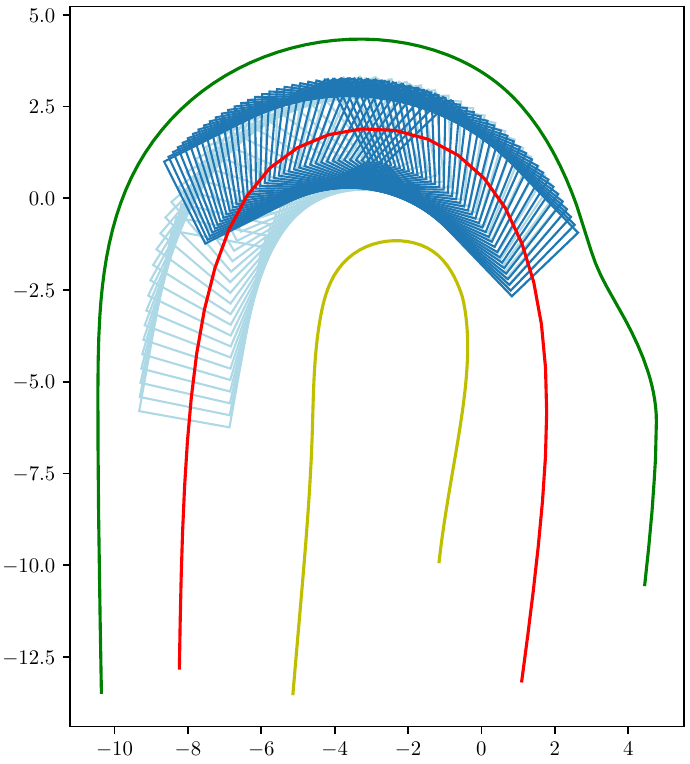}
    \includegraphics[width=0.48\textwidth]{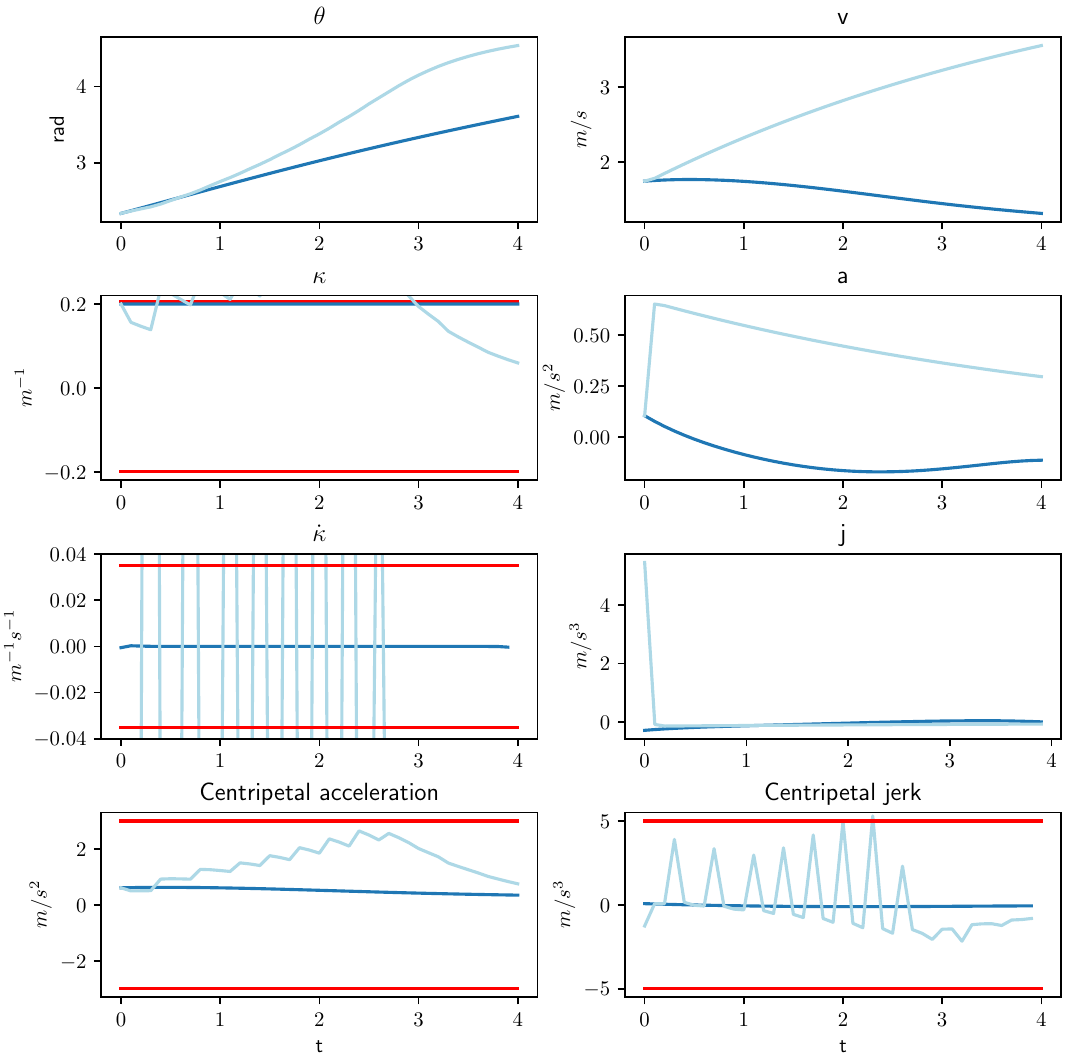}
    \caption{An example extracted from a failed road test planning cycle. The algorithm cannot converge to a feasible trajectory within $50$ ms computation time. We reproduce the problem offline and the algorithm finds the optimal trajectory after $46$ iterations. The results are show in this figure. The planned trajectory is shown in dark blue and the initial guess from P-controller is shown in light blue.}
    \label{fig:failed_uturn}
\end{figure}

We test the proposed algorithm on a list of synthetic problems to analyze the convergence rate (see. Fig.\ref{fig:scenario_straight_road_obs} for example). We also deploy the proposed algorithm on an electric vehicle from Xiaopeng for road tests. The test vehicle equips an Nvidia Drive Orin computation platform. The algorithm is run on a single core of Drive Orin.  The temporal length for the trajectory is $4s$ long. Fig. \ref{fig:computation_time} shows histograms of the computation time and number of optimization iterations from an over one hour test drive on urban roads in Guangzhou, China. We set the cutoff-time for the optimizer to 50 milliseconds. More than 95\% of the planning cycles can get reasonable warm starts from previous planning cycles, and less than 1\% of the cycles fail to return a feasible trajectory. The failure cycles are mostly ($>90\%$) due to improper settings of the drivable corridor, but some cases are indeed difficult to solve within the given cutoff time, see Fig.\ref{fig:failed_uturn} for an example.

\section{Conclusion and Future Work}
We propose a novel nonlinear numerical optimization algorithm for autonomous vehicles. This algorithm is able to reliably find safe and comfortable trajectories for an autonomous vehicle to navigate in cluttered urban driving environments, both in terms of success rate and computation time. 

For future work, we plan to further develop a strategy to better initialize the optimization, make the algorithm more robust to difficult scenarios, and also extend the work to reliably handle dynamic obstacles.  

\bibliography{reference}{}
\bibliographystyle{plain}

\end{document}